\documentclass[letterpaper, 10pt, conference]{ieeeconf}

\IEEEoverridecommandlockouts
\usepackage{algorithm}      
\usepackage{algpseudocode}  
\usepackage{amsmath}
\usepackage{amsfonts}    
\usepackage{graphicx}
\usepackage{multicol}
\usepackage{multirow}
\usepackage{subcaption}
\usepackage{todonotes}
\usepackage{units}
\usepackage[bookmarks=true]{hyperref}
\usepackage{paralist}

\usepackage{newtxtext,newtxmath}
\usepackage{mleftright}
\mleftright
\DeclareMathOperator*{\argmax}{argmax}

\algnewcommand\algorithmicforeach{\textbf{for each}}
\algdef{S}[FOR]{ForEach}[1]{\algorithmicforeach\ #1\ \algorithmicdo}

\floatstyle{ruled}
\newfloat{algorithm}{tbp}{loa}
\providecommand{\algorithmname}{Algorithm}
\floatname{algorithm}{\protect\algorithmname}

\graphicspath{{img/}}

 \title{\LARGE \bf Distributed Data Storage and Fusion
 for Collective Perception \\
 in Resource-Limited Mobile Robot Swarms}

\author{%
  \authorblockN{%
    Nathalie~Majcherczyk,
    Daniel~Jeswin~Nallathambi,
    Tim~Antonelli,
    ~Carlo~Pinciroli}
  \authorblockA{Robotics Engineering, Worcester Polytechnic Institute, MA, USA\\
  Email: \{nmajcherczyk, cpinciroli\}@wpi.edu}
}

\begin{document}

\maketitle

\begin{abstract}
In this paper, we propose an approach to the distributed storage and fusion of data for collective perception in resource-limited robot swarms. We demonstrate our approach in a distributed semantic classification scenario. We consider a team of mobile robots, in which each robot runs a pre-trained classifier of known accuracy to annotate objects in the environment. We provide two main contributions: \emph{(i)} a decentralized, shared data structure for efficient storage and retrieval of the semantic annotations, specifically designed for low-resource mobile robots; and \emph{(ii)} a voting-based, decentralized algorithm to reduce the variance of the calculated annotations in presence of imperfect classification. We discuss theory and implementation of both contributions, and perform an extensive set of realistic simulated experiments to evaluate the performance of our approach.

\end{abstract}

\section{INTRODUCTION}
\label{sec:introduction}

Decentralized collective perception is one of the main activities that a swarm of robots must perform. A major challenge in collective perception is dealing with partial and uncertain knowledge. To deploy robots in safety-critical applications, such as disaster recovery or autonomous driving, researchers strive to increase the reliability of the information used for decision-making. In the context of resource-constrained robot swarms, significant effort has been devoted to solving this problem, with particular focus on the best-of-$n$ formulation \cite{valentini2016},~\cite{ebert2020bayes}.

In this paper, we deal with a collective perception scenario related to, but more complex than, the best-of-$n$ formulation. We assume that individual robots perform observations, from which they individually derive annotations on the environment through a machine learning algorithm. The algorithm, however, might produce incorrect annotations. The aim of our work is to take advantage of the fact that multiple robots can combine individual annotations to derive a set of more accurate consolidated annotations.

\begin{figure}[t]
    \centering
    \includegraphics[width=.7\linewidth]{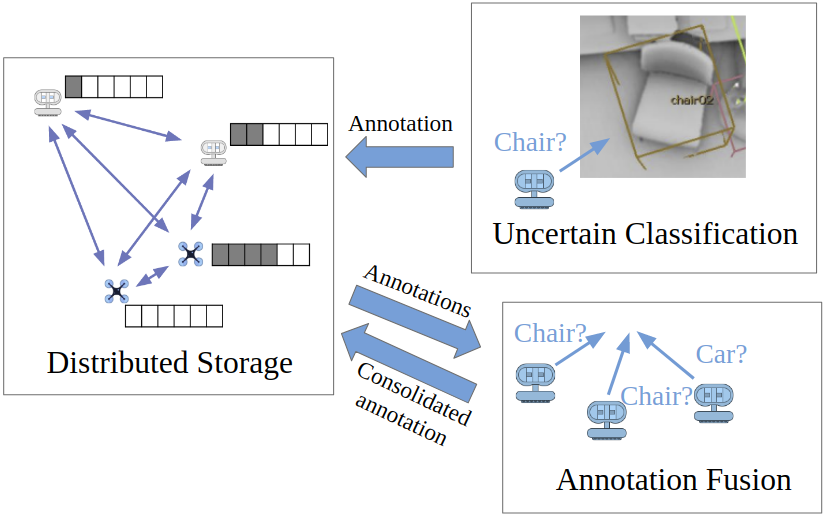}
    \caption{Collective semantic annotation application with annotation consolidation.}
    \label{fig:architecture}
\end{figure}

To showcase our approach, as shown in Fig.~\ref{fig:architecture}, we consider the case in which the robots must collectively construct and store a coherent set of semantic annotations on features found in the environment, such as pieces of furniture, trees, or doorways. To consider low-memory constraints, we assume that the objects are internally stored as bounding boxes, rather than through more detailed representations such as voxels or graphs. These annotated bounding boxes constitute the essence of a distributed \emph{semantic map}~\cite{riazuelo2015roboearth} of the environment. This problem is different from the best-of-$n$ problem. In the latter, the robots must agree on a single choice for the entire swarm; in our problem, the robots must collectively construct a coherent set of annotations.

Semantic mapping poses interesting challenges in the context of resource-constrained collective robotics. Constructing a semantic map entails the collection and storage of large amounts of data. This data is typically processed by the robots individually at first. The result is then shared and fused with the results of other robots to produce a coherent representation of the environment. An effective approach to solve these problems is capable of coping with low memory availability, mobility, noise, uncertainty, and bandwidth limitations~\cite{khodayi2019distributed,atanasov2015decentralized,leung2012decentralized}.

Our approach to semantic mapping draws inspiration from research in ensemble learning~\cite{polikarEnsemble2012}. Ensemble learning reduces the variability of individual models through information fusion~\cite{sunderhauf2018limits}. In ensemble learning, fusion is achieved through bagging and stacking techniques that aggregate models, typically classifiers, by combining their outputs through an averaging process or through a meta-model trained for that purpose~\cite{polikarEnsemble2012,kuncheva2014combining}.

This paper offers two main contributions. First, we present an algorithm to store localized semantic annotations across a swarm of mobile robots, under the form of a shared global memory. Second, we propose an approach to online fusion of uncertain semantic annotations, formalizing the problem and deriving a solution from first principles. We consider a scenario in which the user has already trained a multi-class classifier on an extensive dataset with multiple annotations of various objects \cite{uy2019revisiting}. This classifier is deployed on each robot and used to identify objects for the map, which are shared with the rest of the swarm. The semantic map is constructed by consolidating multiple observations of the same objects into single ones. Through a voting mechanism, our approach copes with the fact that classifiers are imperfect and might mislabel objects. The final result of our approach is a semantic map whose accuracy is superior to that achievable by any individual robot alone. We also provide a closed-form equation to assess the accuracy of a consolidated annotation as a function of the number of votes and the individual classifier accuracy.

Our paper is organized as follows. In Section~\ref{sec:relatedwork} we discuss related work. The components of our framework are described in Section~\ref{sec:methodology}. We report the results of our performance evaluation in Section~\ref{sec:evaluation}, and conclude the paper in Section~\ref{sec:conclusions}.

\section{RELATED WORK}
\label{sec:relatedwork}

\textbf{Best-of-$\mathbf{n}$ problem.}
Previous work in swarm robotics has tackled collective classification problems for one global environmental property. These problems have commonly been referred to as best-of-$n$~\cite{ValFerDor2017}. Valentini \emph{et al.} proposed a honeybee-inspired approach for robots to collectively decide whether a black and white colored environment has a majority of white or black tiles \cite{valentini2016}. Ebert \emph{et al.} applied Bayesian estimation to the same problem with the added explicit group-wide agreement on the output~\cite{ebert2020bayes}. Robust formulations have also been studied in which robots might be affected by noise \cite{crosscombe2017robust} or act in an adversarial manner \cite{mitra2019resilient}. As discussed in the introduction, semantic mapping is a more complex problem, in that it can be seen as a repeated best-of-$n$ problem in which the annotations must also be coherent with each other.

\textbf{Estimation of continuous variables.}
Other work has considered the collective estimation of a continuous value. Early work focused on average consensus~\cite{tahbaz-salehiConsensusRandomNetworks2006,xiaoDistributedAverageConsensus2007,leblancResilientAsymptoticConsensus2013}, in which the robots must agree on the average of individual initial estimates of a quantity of interest. More recently, Albani~\emph{et al.} proposed a collaborative weed mapping application that considers inter-robot networking and motion planning~\cite{albani2017field}. Robots communicate to improve the confidence interval of weed density predictions for each cell in a discrete environment. They assume that robots know the confidence value for each independent prediction and keep the prediction with the maximum confidence. In our work, in contrast, we do not assume the confidence of the prediction known, and resort to a voting process to identify the most likely annotation.

\textbf{Distributed mapping and semantic classification.}
Notable recent work in collective perception include distributed mapping, such as DOOR-SLAM~\cite{lajoie2020door}. In DOOR-SLAM, the robots construct a graph-based map (but not a semantic one) in a decentralized fashion, coping with spurious observations by identifying and rejecting outliers. In distributed semantic classification~\cite{queralta2020collaborative}, effort have concentrated on cloud-based method rather than decentralized ones, such as RoboEarth~\cite{riazuelo2015roboearth}. In the work of Ruiz \emph{et al.}~\cite{ruiz2017building}, a centralized knowledge base is used to build semantic map. Classification uncertainty is solved through Conditional Random Fields, which estimate beliefs on the annotations by combining spatial and contextual knowledge.

\section{PROBLEM STATEMENT}
\label{sec:probstat}

In this section, we explain our main modelling assumptions and formulate the distributed storage and collective annotation problems.

\subsection{Assumptions}

\subsubsection{Multi-Robot System} 

We consider a system of $N$ robots, each with a unique identifier $i \in \{1,2,\ldots,N\}$. Each robot can exchange messages within a communication range $C$ with its neighbors $\mathcal{N}_i(t)$. Outgoing messages are limited in size at each time step $t$ to model a finite broadcasting bandwidth. Furthermore, each robot has a limited memory capacity $M_i$ devoted to the housekeeping of a shared data structure. This capacity is subdivided into storage capacity $S_i$ and routing capacity $R_i$. The available memory of a robot at a given time is denoted by $m_{i}(t)$. Robots move according to a diffusion policy with obstacle avoidance~\cite{diffusion}. We assume that each robot can localize itself in a global frame. We further assume robots to be equipped with a perception module. To focus on the problems of storage and fusion, in this paper we abstract away the inner workings of the module and simulate its behavior through the following assumptions: \begin{inparaenum}[\it (i)]
\item the perception module uses a sensor with physical specifications constraining its viewing frustum (see Figure~\ref{fig:object_detection});
\item the perception module is able to determine the center position of objects perfectly;
\item the perception module annotates objects imperfectly (see Section~\ref{sssec:simulated_classifier}).
\end{inparaenum}

\subsubsection{Object Representation and Detection} 

\begin{figure}[t]
    \centering
    \includegraphics[width=0.5\linewidth]{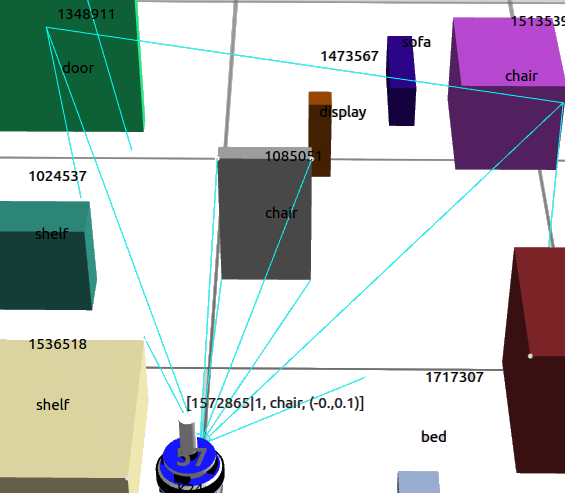}
    \caption{The front corners of an object are inside the sensor viewing frustum.}
    \label{fig:object_detection}
\end{figure}

We consider robots to be in an environment with non-uniformly distributed objects of different types. We use 3D bounding boxes to represent these objects in space. A bounding box is fully described by its center, dimensions, and orientation~\cite{mousavian20173d}. We assume that the sensor of the perception module can detect and localize objects when their front face corners fall inside the sensor viewing frustum (see Figure~\ref{fig:object_detection}). 

\subsubsection{Simulated Classifier} 
\label{sssec:simulated_classifier}
Within the perception module, we simulate the statistical behavior of a trained multi-class object classifier. We leverage the work of Carillo \textit{et al.}~\cite{carrillo2014probabilistic} deriving the posterior distribution of the balanced accuracy, conditioned on predictions of a trained classifier on a sample test dataset. We use this posterior distribution to generate either correct or incorrect object annotations upon each object detection. To distinguish cases of erroneous labels, we further modelled the leftover probability distributions using the confusion matrices for the classifier. The statistical data used for these probabilistic models comes from the SceneNN benchmarking data~\cite{scenenn-3dv16}.
SceneNN provides raw outcomes and accuracies for several multi-class classifiers tested extensively on a catalog of point clouds. We use the PB-T50-RS dataset with translation, rotation and scaling perturbations of the bounding boxes to compute the posterior distribution modelling the classifier statistical behavior.

\subsection{Distributed Storage Problem Formulation}
\label{ssec:distributed_storage}

The first goal of our work is to distribute object annotation data across robots in a way that makes efficient use of the collective memory. Each robot can hold a defined number of data items specified by its memory capacity $M_i$.
Besides memory constraints, we want to account for communication-related costs.

Since we want to pack data items into bins (robots) with different properties, we formulate the distributed storage problem as a heterogeneous bin packing optimization problem. In particular, we consider the variant known as the Variable Cost and Size Bin Packing Problem (VCSBPP)~\cite{crainic2011efficient}. 

Given a set of items $\mathcal{T}$ with different volumes $v_{\tau}$ to be loaded into a set of available bins $\mathcal{I}$ with different capacities $M_i$ and costs $c_i$, we aim to minimize the total cost of the bins selected to store items:
\begin{subequations}
\begin{align}
\min_{a, s}  \quad & f(a, s) = \sum_{i \in \mathcal{I}} c_i s_i  \label{seq:cost} \\
\textrm{s.t.} \quad & \sum_{i \in \mathcal{I}} a_{\tau i} = 1 \quad & \forall \tau \in \mathcal{T} \label{seq:tuple_once}\\
  & \sum_{\tau \in \mathcal{T}} v_{\tau} a_{\tau i} \leq M_{i}s_i  \quad & \forall i \in \mathcal{I} \label{seq:capacity_constraint}\\
  & a_{\tau i} \in \{0,1\} \quad & \forall \tau \in \mathcal{T},\ \forall i \in \mathcal{I} \\
  & s_{i} \in \{0,1\} \quad & \forall i \in \mathcal{I} 
\end{align}
\label{eq:formulation}
\end{subequations}

\noindent with $ \{ s_i \}_{i \in \mathcal{I}}$ denoting bin-selection variables, and $s_i$ equalling 1 when a bin $i$ is selected for storing items. The set $\{ a_{\tau i} \}_{t \in \mathcal{T}, i \in \mathcal{I}}$ contains item-to-bin assignment variables; $a_{\tau i}$ equals 1 if item $\tau$ is assigned to bin $i$. In our variant of the problem, we define bin costs $c_i$ as follows:
\begin{equation}
    \label{eq:costs}
    c_i = \cfrac{1}{|\mathcal{N}_{i}| \cdot m_i} = \cfrac{1}{|\mathcal{N}_{i}| \cdot (M_i - \sum_{\tau \in \mathcal{T}} v_{\tau} a_{\tau i}) }
\end{equation}
We select bin cost to be inversely proportional to the product of the number of neighbors and the memory left after the assignment. This is to model practical considerations such as potential for disconnections and subsequent memory overflows, which are not otherwise modelled in the VCSBPP.
%

\subsection{Annotation Consolidation Problem Statement}
\label{ssec:collective_annotation}

Our second goal is to improve the map annotation accuracy by using at least $V$ votes for each object to identify in the environment. We seek to derive the ensemble probability of success in predicting the correct class  $p_{ens}(V,\, p)$ as a non-decreasing function of $V$ that is always at least as large as the accuracy $p$ of the correct class.

\section{METHODOLOGY}
\label{sec:methodology}

\subsection{Overview}

Figure~\ref{fig:architecture} shows the interconnection between the different components of the collective semantic annotation application with label consolidation. Robots move in an environment with objects to annotate. Upon detecting an object and deciding to record an annotation, each robot assigns an uncertain class annotation for the location. The robots writes the annotation into the shared memory as described in Section~\ref{ssec:swarmmesh}. Upon storing multiple annotations from the same location, robots query the shared memory for all data from that location to retrieve all the related annotations. If they receive enough annotations back from the query, the robots write the most frequent label for that location into the shared memory as a \textit{consolidated annotation}. The exact consolidation mechanism and the resulting ensemble accuracy are formalized in Section~\ref{ssec:plurality_vote}. Over time, as robots explore the environment, the shared memory stores an increasingly more complete map of the annotated objects in the form of consolidated annotations. Section~\ref{ssec:local_routine} explains the local routine run by the robots to perform the various actions across components.

\subsection{Distributed Storage Through SwarmMesh}
\label{ssec:swarmmesh}

We implement a shared memory to enable distributed storage across the robots. For this purpose we used SwarmMesh~\cite{majcherczyk2019swarmmesh}, a decentralized data structure for mobile swarms. As part of the work in this paper, we also provide a generic, open-source C++ implementation of this data structure.\footnote{\url{https://github.com/NESTLab/SwarmMeshLibrary}} SwarmMesh is based on a distributed online heuristic solution to the Bin Packing problem formulated in Section~\ref{ssec:distributed_storage}, Eqs.~(\ref{eq:formulation})-(\ref{eq:costs}).

SwarmMesh stores data in the form of tuples (key-value pairs).
The tuple key \texttt{k} is used for both storing and addressing tuples. It consists of two parts:
\begin{inparaenum}[\it (i)]
\item a unique tuple identifier $\tau$ and
\item a tuple hash $\rho_{\tau}$ that depends on the tuple value \texttt{v} to be stored.
\end{inparaenum}
When a robot $i$ creates a new tuple, the unique tuple identifier $\tau$ is calculated as the concatenation of the binary representations of $i$ and of the count of tuples created so far by the robot. The tuple value \texttt{v} contains the annotation $\lambda_{\nu}$, the box center, the box orientation and the vector from the center of the box to the front right corner of the box. These variables fully describe an annotated 3D bounding box. 

Each robot calculates a so-called NodeID, $\delta_i$, which determines which tuples can be stored on a robot and which cannot. 
The robots compute their NodeIDs at each time step as follows:
\begin{equation}
\delta_i(t) =
\begin{cases}
 m_i(t) \cdot |\mathcal{N}_i(t)| & \text{if } |\mathcal{N}_i(t)| > 0\\
 1 & \text{otherwise} 
\end{cases}
\label{eq:nodeid}
\end{equation}
Intuitively, the NodeID encodes how willing a robot is to store a new tuple, but also how desirable it is to store a tuple on that robot. In Eq.~\eqref{eq:nodeid}, the NodeID increases linearly with the amount of available memory (more room for tuples) and the number of neighbors of a robot (better connectivity, lower likelihood of disconnections, better routing). As the robots move and store or delete tuples, the NodeIDs change over time.

The NodeIDs form a distribution that defines an hierarchy among robots. To decide which robot eventually stores a tuple, SwarmMesh has an analogous mechanism that imposes an hierarchy among tuples. A tuple is passed from robot to robot until $\delta_i(t) > \rho_{\tau}$. Intuitively, we want tuples with a high hash $\rho_\tau$ to occupy the best robots in the network. Our insight is that the \emph{most uncertain} annotations are those that should be mapped to higher values of $\rho_\tau$, because this makes them easier to query and route. This makes it more efficient to consolidate uncertain annotations into annotations with a sufficient level of confidence. In contrast, annotations with high levels of certainty need a less critical placement. To formalize this notion, we calculate $\rho_\tau = R(\lambda_\nu)$, where $R(\cdot)$ is a staircase function increasing with the uncertainty of annotation $\lambda_\nu$.


SwarmMesh offers a wide API of operations to handle tuples. In this work, we use the following operations: 
\begin{itemize}
\item\texttt{store(k,v)} which assigns a tuple to a suitable robot in the data structure; 
\item \texttt{get(x,y,r)}, which returns all the tuples located within a radius $r$ of the point $(x,y)$ expressed in a global reference frame; 
\item \texttt{erase\_except\_tuple(x,y,r,k)}, which removes from the data structure all tuples located within a radius $r$ around the point $(x,y)$ in the global reference frame, except for the tuples with unique identifier $\tau$ in the list \texttt{k}.
\end{itemize}

\subsection{Annotation Consolidation Through Plurality Voting}
\label{ssec:plurality_vote}

\textbf{Plurality vote.}
Upon querying the shared memory for a particular location in space, the querying robot receives class annotations predicted by various robots. If the number of votes $n$ is greater than or equal to the minimum number of votes $V$, the robot aggregates these votes into a consolidated annotation $\Bar{\lambda}$ for the object through plurality voting:
$$ 
\Bar{\lambda} = \argmax_{\text{classes}} \sum_{\nu=1}^{n} I(\lambda_{\nu} = \text{class}) 
$$ 
where $I$ is the indicator function. In case of ties, the consolidated annotation is selected at random from the tied options.

\textbf{Ensemble Probability of Success.}
We seek to derive the probability of success for $n$ independent votes, each for one of $c$ classes where $c \ge n$. We assume that class 1 is correct and that each vote is for class 1 with probability $p$. The probability of an incorrect vote is then $1-p$, and we make the simplifying assumption that each incorrect class is equally likely. 
We denote the total vote count for each class as a vector $z = (z_1, z_2, \ldots, z_c)$ where $z_i \in \{0,1,\ldots,n\}$ and $z_1 + z_2 + \cdots + z_c = n$. This vector follows a multinomial distribution with probability mass function
\[\phi(z\,|\,n,p) = \binom{n}{z_1,\ldots,z_c}p^{z_1}\left(\frac{1-p}{c-1}\right)^{n-{z_1}}\]
\noindent where

\[\binom{n}{z_1,\ldots,z_c} = \frac{n!}{z_1!\cdots z_c!}\]

\noindent denotes the multinomial coefficient.

Without knowing the correct class, we can estimate it from an observed vector as the class with the most votes, whether a majority or plurality. In the case where several classes are tied with the highest number of votes, we select one of those classes at random. Again considering that $z_1$ represents the true correct class, we use $z^*$ to denote a vector of vote counts for which $z_1 \ge z_i$, for all $i = 2, 3, \ldots, c$. The probability of identifying the correct class with $n$ total votes is then
\begin{equation}p_{ens}(n, p) = \sum_{z^*} \phi(z^*\,|\,n,p)P(\mathrm{success}\,|\,z^*)\label{eq:ensemble_one}\end{equation}
since all other vectors $z$ result in a success probability of zero. Because the largest term in $z^*$ appears first and all incorrect classes are chosen with equal probability, we can express \eqref{eq:ensemble_one} using integer partitions.

\textbf{Integer partitions.}
An integer partition $\xi$ is a nonincreasing sequence of positive integers $\xi_1 \ge \xi_2 \ge \cdots \ge \xi_\omega$. We say that $\xi$ is a partition of $n$, denoted $\xi \vdash n$, if $\sum_{i=1}^\omega \xi_i = n$, and we consider the set of all such $\xi$ to be $\mathcal{P}$.  The $\xi_i$ are called \textit{parts} of the partition, and we define the \textit{length} of the partition $\ell(\xi) := \omega$. An alternative formulation is to consider the infinite sequence of multiplicities of each part $\xi = (k_1, k_2, \ldots)$ where $k_j \in \{0,1,\ldots\}.$ Here, $\xi \vdash n$ if $\sum_{j=1}^\infty jk_j = n$, and the length $\ell(\xi) = \sum_{j=1}^\infty k_j = \omega$, since all but a finite number of $k_j$ are zero \cite{andrews1998theory}. For example, $\xi = (5,3,2,1,1)$ in part notation and $\xi = (2,1,1,0,1,0,\ldots)$ in multiplicity notation represent the same partition of $n=12$. We will refer to both the parts $\xi_i$ and multiplicities $k_j$ of the same partition throughout.

Note that we can map each vector $z^*$ to an integer partition $g(x^*) = \xi$, where $\xi_1$ represents the number of votes for the correct class, and the other $\xi_i$ represent the positive numbers of votes for $\omega-1$ incorrect classes. Since all incorrect classes are equally likely, each $z^*$ that maps to a particular $\xi$ yields the same probability of success. Thus,
\begin{align}p_{ens}(n, p) &= \sum_{\substack{\xi \vdash n\\\xi \in \mathcal P}} \sum_{\substack{z^*:g(z^*)=\xi}} \phi(z^*\,|\,n,p)P(\mathrm{success}\,|\,z^*)\notag\\
&= \sum_{\substack{\xi \vdash n\\\xi \in \mathcal P}} \left| g^{-1}\big(\{\xi\}\big)\right|\phi(\xi\,|\,n,p)P(\mathrm{success}\,|\,\xi)\label{eq:ensemble_form}
\end{align}
where $\left|g^{-1}\big(\{\xi\}\big)\right|$ is the cardinality of the preimage of $\{\xi\}$, i.e., the number of $z^*$ that map to $\xi$.

To determine $\left|g^{-1}\big(\{\xi\}\big)\right|$, we start with the $\binom{c-1}{\omega-1}$ combinations of incorrect classes and count the ways to assign vote counts with multiplicities $k_j$ to each. There are 
\[\binom{\omega-1}{k_1,k_2,\ldots,k_{\xi_1}-1}\]
such assignments, where the $k_{\xi_1}-1$ derives from the fact that one of the spots for the largest possible vote count is taken by the correct class. Together, we have
\[\left|g^{-1}\big(\{\xi\}\big)\right| = \binom{c-1}{\omega-1}\binom{\omega-1}{k_1,k_2,\ldots,k_{\xi_1}-1}\]

For the conditional probability $P(\mathrm{success}\,|\,\xi)$, when $\xi_1$ is the strictly largest part, it is chosen with probability 1; otherwise, there is a $k_{\xi_1}$-way tie, from which a class is chosen at random. In both cases, the correct class is chosen with probability $P(\mathrm{success}\,|\,\xi) = 1/k_{\xi_1}$. Substituting into \eqref{eq:ensemble_form} and simplifying gives
\begin{multline}
p_{ens}(n, p) = \\
\frac{1}{c} \sum_{\substack{\xi \vdash n\\\xi \in \mathcal P}} \binom{n}{\xi_1,\ldots,\xi_\omega}\binom{c}{\omega}\binom{\omega}{k_1,\ldots,k_{\xi_1}}p^{\xi_1}\left(\frac{1-p}{c-1}\right)^{n-\xi_1}
\label{eq:ensemble_accuracy}
\end{multline}

\subsection{Robot Local Routine}
\label{ssec:local_routine}

Robots each run a local procedure in a loop. We refer to this procedure as the \textit{control step} and outline it from the perspective of robot $i$ in Algorithm~\ref{alg:control_step}. Every time step, the robot performs one iteration of this loop. 

At the beginning of the procedure, the robot receives and deserializes messages from its neighbors. Certain messages are meant for specific recipients only (point-to-point) and get discarded by any other receiving robot during deserialization.  
The robot then computes and performs a motion increment according to a diffusion policy~\cite{diffusion}.

Next, the robot checks the output of its perception module if it is not in recording timeout. The recording timeout is a delay imposed to avoid making successive observations of the same object. If the perception module has detected an object, the robot creates a tuple and starts a \texttt{store()} operation on the shared memory. The payload of the tuple contains the object annotation and the bounding box location and dimensions.

If the robot is not in a querying timeout, it reviews the tuples stored in its memory by bounding box location. If the robot finds multiple tuples with the same location in its own local memory, it starts a \texttt{get()} query on SwarmMesh to retrieve all the tuples currently stored with that bounding box location. It then saves the meta data for the started query.

The robot goes through the queries it has started and checks whether it has not received replies for a period longer than a time threshold. If the number of tuples is larger than the minimum number of votes, the robot creates and writes a tuple into SwarmMesh with the consolidated annotation. It also start an \texttt{erase\_except\_tuple()} request of all tuples with the given location except for the newly created tuple.

SwarmMesh routing at line~\ref{algline:routing} of Alg.~\ref{alg:control_step} decides which requests and replies need to be routed and to which robot. A robot may have to send a reply to a request and continue propagating the request or route a neighbor's reply or discard the request or reply. The details of the routing process are reported in~\cite{majcherczyk2019swarmmesh}.

\begin{algorithm}[h]
\caption{Control step - message $\mu$, neighbors $\mathcal{N}_i$,
requests $\mathcal{Q}$, replies $\mathcal{R}$}
\label{alg:control_step}
\begin{scriptsize}
\begin{algorithmic}[1]
\Procedure{Control\_step}{}
\State$\mathbf{receive}\,\,\{ \mu_j \}_{j \in \mathcal{N}_{i}}$ \Comment{RX from neighbors}
\ForEach{$j \in \mathcal{N}_i$}
\State $(j, \delta_j, \mathcal{Q}_j, \mathcal{R}_j ) \gets \mu_j$ \Comment{Deserialization}
\EndFor
\State$\mathbf{move()}$ \Comment{Diffusion motion}
\If{not in recording timeout} 
    \State \texttt{v} $\gets \mathbf{senseObject()}$
    \If{object detected}
    \State start recording timeout
    \State \texttt{k} $\gets \mathbf{hash(\texttt{v})} $
    \State $\mathcal{Q}_i \gets \mathcal{Q}_i\, \cup $ \texttt{store(k, v)} \Comment{Write label}
    \EndIf
\EndIf 
\If{not in querying timeout} 
    \If{holds multiple tuples from same \texttt{(x,y)}}
        \State start querying timeout
        \State $\mathcal{Q}_i \gets \mathcal{Q}_i\, \cup $ \texttt{get(x, y, 0)} 
        \Comment{Ask for labels}
        \State save query information
    \EndIf
\EndIf
\ForEach{started query}
    \If{results ready and enough votes}
        \State \texttt{v} $\gets$ plurality vote \Comment{Consolidated label}
        \State \texttt{k} $\gets \mathbf{hash(\texttt{v})}$
        \State $\mathcal{Q}_i \gets \mathcal{Q}_i\, \cup $ \texttt{store(k,v)} 
        \State $\mathcal{Q}_i \gets \mathcal{Q}_i\, \cup $ \texttt{erase\_expt\_tuple(x,y,0,k)}
    \EndIf
\EndFor
\State $(\mathcal{\bar{Q}}_i, \mathcal{\bar{R}}_i) \gets \mathbf{route(\mathcal{Q}_{\bullet}, \mathcal{R}_{\bullet})}$ \Comment{SwarmMesh routing} \label{algline:routing}
\State $\mu_i \gets (i, \delta_i, \mathcal{\bar{Q}}_i, \bar{\mathcal{R}}_i)$ \Comment{Serialization}
\State $\mathbf{send}\left( \mu_i \right)$ \Comment{TX to neighbors}
\EndProcedure
\end{algorithmic} 
\end{scriptsize}
\end{algorithm}

\section{EVALUATION}
\label{sec:evaluation}

We performed the evaluation of our approach using the physics-based ARGoS multi-robot simulator~\cite{Pinciroli:2012} with data from the SceneNN benchmark~\cite{scenenn-3dv16}. The main parameters of interest are the minimum number of votes $V$ and the number of robots $N$. The former relates to the ensemble accuracy of the map through Eq.~\eqref{eq:ensemble_accuracy} and Figure~\ref{fig:ensemble_accuracy}. The latter changes the scale of the distributed system and the degree of parallelization in collecting data. 

\subsection{Simulation Parameters}

\textbf{Robots and environment.} 
We study the influence of the number of robots $N$ by running experiments with 30, 60 and 90 robots. The robots have a communication range $C$ of \unit[2]{m} and move at a forward speed of \unit[5]{cm/s}. The robots diffuse and avoid obstacles in an $\unit[8\times8]{m^2}$ environment. 
Given these settings, the robot density is such that robots are within communication range most of the time, but a significant number of intermittent disconnections occur. We do not consider line-of-sight obstructions.

\begin{figure}[t]
    \centering
    \includegraphics[width=.7\linewidth]{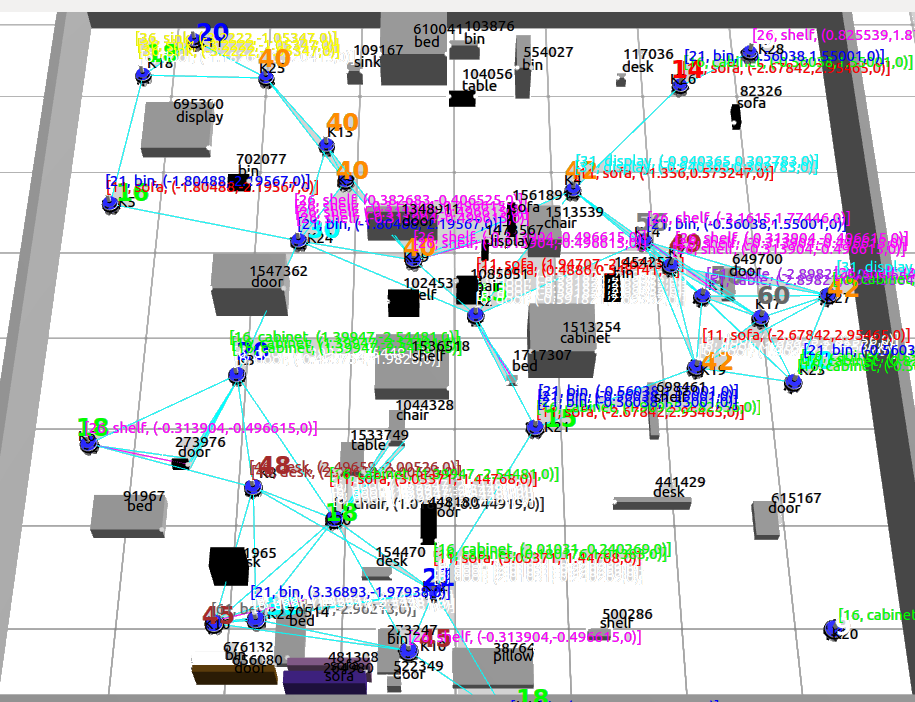}
    \caption{Storing map annotations collectively and reducing uncertainty through data fusion.}
    \label{fig:concept}
\end{figure}

In order to run experiments representative of collective annotation in an indoors environment, we import 40 objects of 13 types from the SceneNN dataset. Therefore, the objects are non-uniformly distributed in space. We made the following adjustments when importing scene 005 of the dataset: \begin{inparaenum}[\it (i)]\item we limited physical dimensions to fit in the viewing frustum;
\item we lowered objects to the ground level;
\item we rotated them to face towards the center of each ``room'';
\item we removed overlapping objects;
\item we re-labelled ``unknown'' objects by drawing from the list of object classes at random.
\end{inparaenum}
Figure~\ref{fig:concept} shows the imported environment with these modifications.


\textbf{Classifier.} We simulate the statistical behavior of the BGA-DGCNN classifier using data from the SceneNN dataset~\cite{scenenn-3dv16,uy2019revisiting}. The procedure to generate a posterior distribution is described in Section~\ref{sssec:simulated_classifier}. 
Each observed annotation $\lambda_{\nu}$ in the simulation is given by drawing from this distribution.

\begin{figure}[b]
    \centering
    \includegraphics[width=0.8\linewidth]{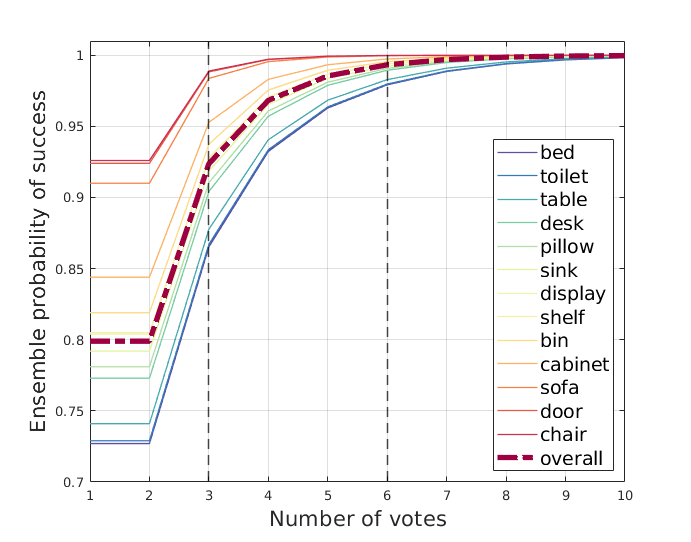}
    \caption{Ensemble probability of success per class of object. This graph is based on the per-class accuracies in Table~\ref{tab:classifier_accuracy}.}
    \label{fig:ensemble_accuracy}
\end{figure}

\textbf{Ensemble accuracy.}
In order to set $V$ according to a desired range of probabilistic accuracy, we compute the ensemble accuracy $p_{ens}(n, p)$ for each class of objects for a given number of votes $n$ using Equation~\ref{eq:ensemble_accuracy}. We use the per-class accuracy $p$ for BGA-DGCNN shown in Table~\ref{tab:classifier_accuracy}, as reported in~\cite{uy2019revisiting}. Figure~\ref{fig:ensemble_accuracy} is the resulting curve. In our experiments, we vary the threshold $V$ in the range between the two vertical lines in Figure~\ref{fig:ensemble_accuracy}.

\begin{table}[t]
\centering
\caption{BGA-DGCNN classifier class accuracy on SceneNN PB-T50-RS dataset.~\cite{uy2019revisiting}}
\label{tab:classifier_accuracy}
\begin{scriptsize}
\begin{tabular}{|l|l|l|l|l|l|l|} 
\hline
\textbf{bin} & \textbf{cabinet} & \textbf{chair} & \textbf{desk} & \textbf{display} & \textbf{door} & \textbf{shelf} \\
\hline
\multicolumn{1}{|c}{81.9} & \multicolumn{1}{|c}{84.4} & \multicolumn{1}{|c}{92.6} & \multicolumn{1}{|c}{77.3} & \multicolumn{1}{|c}{80.4} & \multicolumn{1}{|c}{92.4} & \multicolumn{1}{|c|}{80.5} \\
\hline
\textbf{table} & \textbf{bed} & \textbf{pillow} & \textbf{sink} & \textbf{sofa} & \textbf{toilet} & \textbf{overall}\\ 
\hline
\multicolumn{1}{|c|}{74.1} & \multicolumn{1}{c|}{72.7} & \multicolumn{1}{c|}{78.1} & \multicolumn{1}{c|}{79.2} & \multicolumn{1}{c|}{91} & \multicolumn{1}{c|}{79.7} & \multicolumn{1}{c|}{75.7} \\
\hline
\end{tabular}
\end{scriptsize}
\end{table}

\textbf{SwarmMesh parameters.} We select the data structure parameters so as to maintain a reasonable load factor, i.e., the ratio of data to store over the available storage capacity. 
We set the memory capacity of robots $M_i$ to 20 tuples with a target of 10 tuple for storage and an allowance of 10 for the routing queue. We set the step size in the hashing function $R(\lambda_\nu)$ to 5 (see Section \ref{ssec:swarmmesh}).


\subsection{Mapping Performance}

\begin{figure}[t]
    \centering
    \includegraphics[width=.8\linewidth]{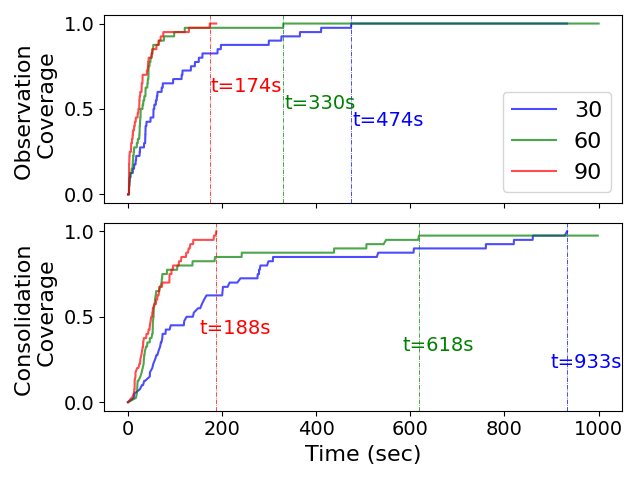}
    \caption{Map coverage over time for $V=3$ and $N \in \{30, 60, 90\}$. Cursors show the time when the maximum coverage was first reached.}
    \label{fig:coverage_across_N}
\end{figure}

\textbf{Effect of the number of robots.} 
Using more robots distributes sensing spatially and leads to a more densely connected robotic network. 
As a result, we expect that increasing the number of robots $N$ speeds up the collection and aggregation of data needed to complete the semantic map. In our setting, map coverage refers to the ratio of \textit{covered} objects to the total number of objects. We make a distinction between two types of coverage: \begin{inparaenum}[\it (i)] 
\item the \textit{observation coverage} which considers an object covered if at least one robot annotated the object;
\item the \textit{consolidation coverage} which considers an object covered if there exists a consolidated annotation $\bar{\lambda}$ for the object in the shared memory. 
\end{inparaenum}
Figure~\ref{fig:coverage_across_N} shows the temporal curve of both types of coverage across different values of $N$ and a set threshold of votes $V=3$.  The curves of map coverage over time increase faster with a higher number of robots. The consolidation coverage rises slower than the observation coverage since it depends on the consolidation process that queries the shared memory as described in Section~\ref{ssec:local_routine}. With $N=90$, the delay for consolidating all the annotations is \unit[14]{s} from the time all objects have been observed.

\begin{figure}[t]
    \centering
    \includegraphics[width=.8\linewidth]{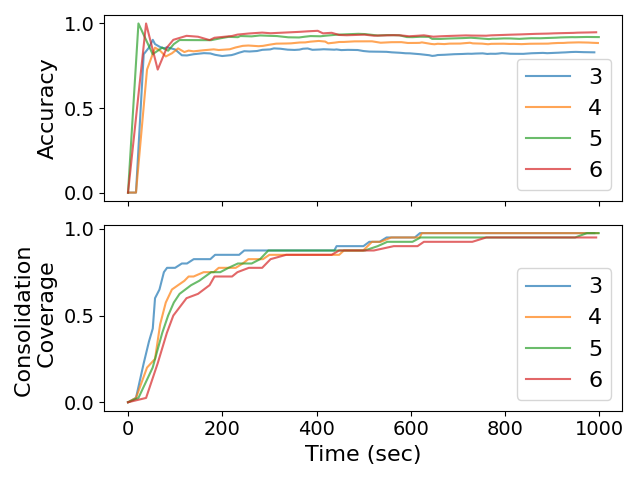}
    \caption{Map accuracy and coverage over time for $N=60$ and $V \in \{3, 4, 5, 6\}$.}
    \label{fig:accuracy_coverage_N60}
\end{figure}

\textbf{Effect of the number of votes.} 
The observation coverage is independent of $V$. Figure~\ref{fig:accuracy_coverage_N60} shows the map accuracy and the consolidation coverage over time across values of $V$. The experimental map accuracy is the ratio of correct consolidated annotations to the number of consolidated annotations. We observe that the map accuracy increases with $V$ at the cost of a slower coverage. 



\subsection{Memory-Related Performance}

\begin{figure}[t]
    \centering
    \includegraphics[width=.8\linewidth]{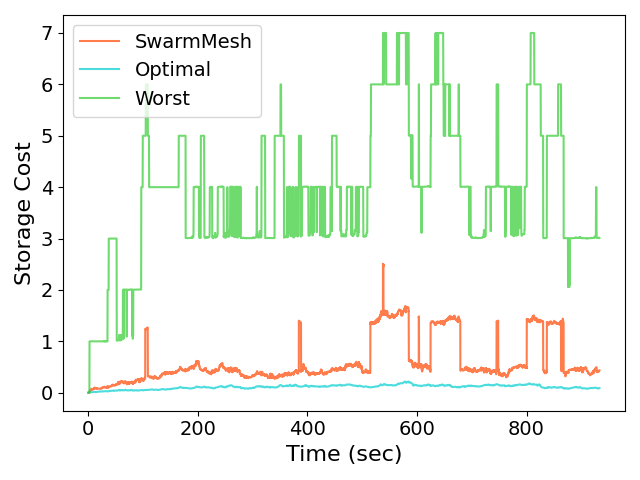}
    \caption{VSCBPP cost over simulation time for $V=3$ and $N=30$.}
    \label{fig:VSCBPP_cost}
\end{figure}

\textbf{Distributed storage cost.} Figure~\ref{fig:VSCBPP_cost} shows the total storage cost (\ref{seq:cost}) of the tuples-to-robots assignment realized in practice over time in the simulation, and optimal and worst case curves. At every time step, the tuple set $\mathcal{T}$ and the number of neighbors of each robot $|\mathcal{N}_i|$ are updated. In our evaluation, we consider tuples of unit volume ($v_\tau = 1,\, \forall\tau$) and an homogeneous memory capacity across robots ($M_i = M, \, \forall i$). We calculate the cost obtained in simulation using the heuristic tuple-to-robot assignment rule of \textit{SwarmMesh}. In order to tell apart cases in which tuples are assigned to robots with null $|\mathcal{N}_i|$ and/or $m_i$, we cap each term in the sum to 1. These cases result in jumps of 1 of the cost in Figure~\ref{fig:VSCBPP_cost}. 

The VSCBPP problem is NP-hard but our experimental settings ($v_\tau = 1,\, \forall\tau$) and ($M_i = M, \, \forall i$) allow for optimally solving instances of the simulation with $N=30$, $V=3$. We reduce the space of the exhaustive search by noting that permutations of the rows of $(a_{\tau i})$ lead to equivalent solutions in terms of cost. We enumerate integer partitions of $|\mathcal{T}|$ with fewer than $|\mathcal{I}|$ parts and such that all parts are smaller than $M$. For each such partition, we determine the minimal cost using the rearrangement inequality for the product of $1/|\mathcal{N}_i|$ and $1/m_i$:
$$
\sum_{i \in \mathcal{I}} \cfrac{1}{|\mathcal{N}_i|} \cdot \cfrac{1}{m_i} \geq \cfrac{1}{|\mathcal{N}_{\sigma(1)}|} \cdot \cfrac{1}{|m_{\pi(1)}|}\, + \dots + \cfrac{1}{|\mathcal{N}_{\sigma(|\mathcal{I}|)}|} \cdot \cfrac{1}{|m_{\pi(|\mathcal{I}|)}|}
$$
where $\sigma(.)$ is an ascending ordering of $|\mathcal{N}_i|$, and $\pi(.)$ is a descending ordering of $m_i$.

Given $|\mathcal{N}_i|$ and $\mathcal{T}$, we compute the worst cost by assigning one tuple to each robot with $|\mathcal{N}_i| = 0$, filling the memory ($m_i = 0$) of as many robots as possible given the number of tuples $|\mathcal{T}|$ and placing the remainder of tuples in the robot of lowest non-zero $|\mathcal{N}_i|$.


\begin{figure}[t]
    \centering
    \includegraphics[width=.8\linewidth]{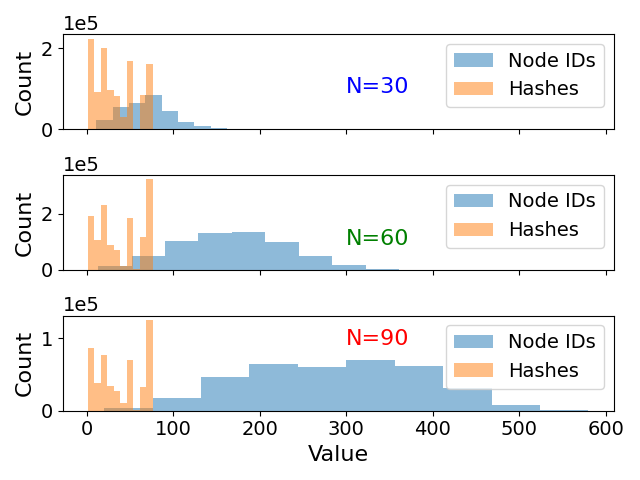}
    \caption{Histograms of data hashes and NodeIDs for $V = 6$ and $N \in \{30, 60, 90\}$.}
    \label{fig:sm_distributions}
\end{figure}

\textbf{SwarmMesh dimensioning.} Figure~\ref{fig:sm_distributions} validates our selection of the \textit{SwarmMesh} parameters. It shows that the bias of the NodeID distribution is greater than that of the distribution of tuple hashes. This indicates that our insight that uncertain annotations should correspond to higher hash values is appropriate, i.e., it is likely to find a robot of greater NodeID than a given tuple hash. We keep the parameters constant across numbers of robots. This means that with higher swarm size $N$, the collective memory capacity $N\cdot M$ is over-sized and the key partitioning matches tuples to robots randomly. In practice, this means that the robot writing the tuple typically keeps it locally, which leads to reduced communication overhead at greater swarm sizes.

\subsection{Communication Load}

\begin{figure}[t]
    \centering
    \includegraphics[width=.8\linewidth]{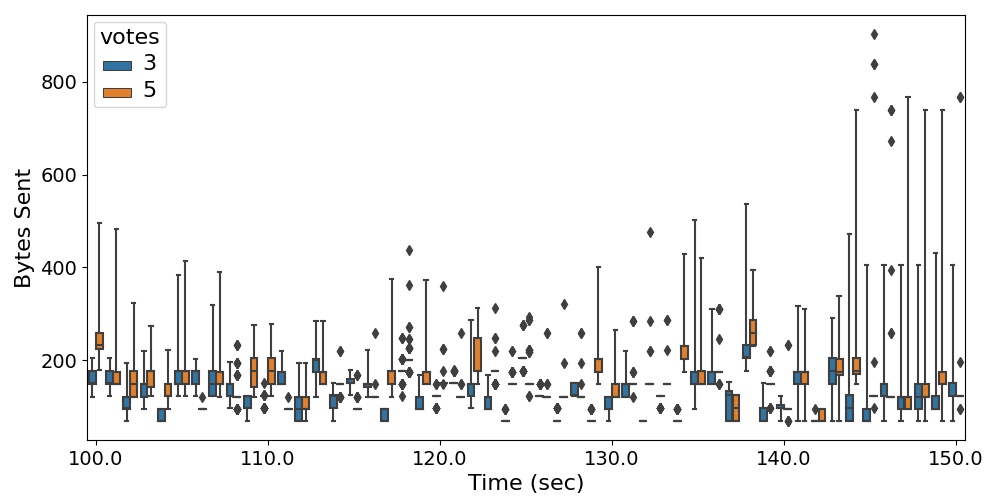}
    \caption{Bytes sent per second per robot over time for $N=60$ and $V \in \{3, 5\}$.}
    \label{fig:bandwidthV}
\end{figure}

\textbf{Effect of the number of votes.} 
We find that the number of bytes sent increases with the minimum number of votes $V$ (Figure~\ref{fig:bandwidthV}). This indicates that an increase in map accuracy comes at the cost of a higher communication overhead. 

\begin{figure}[t]
    \centering
    \includegraphics[width=.8\linewidth]{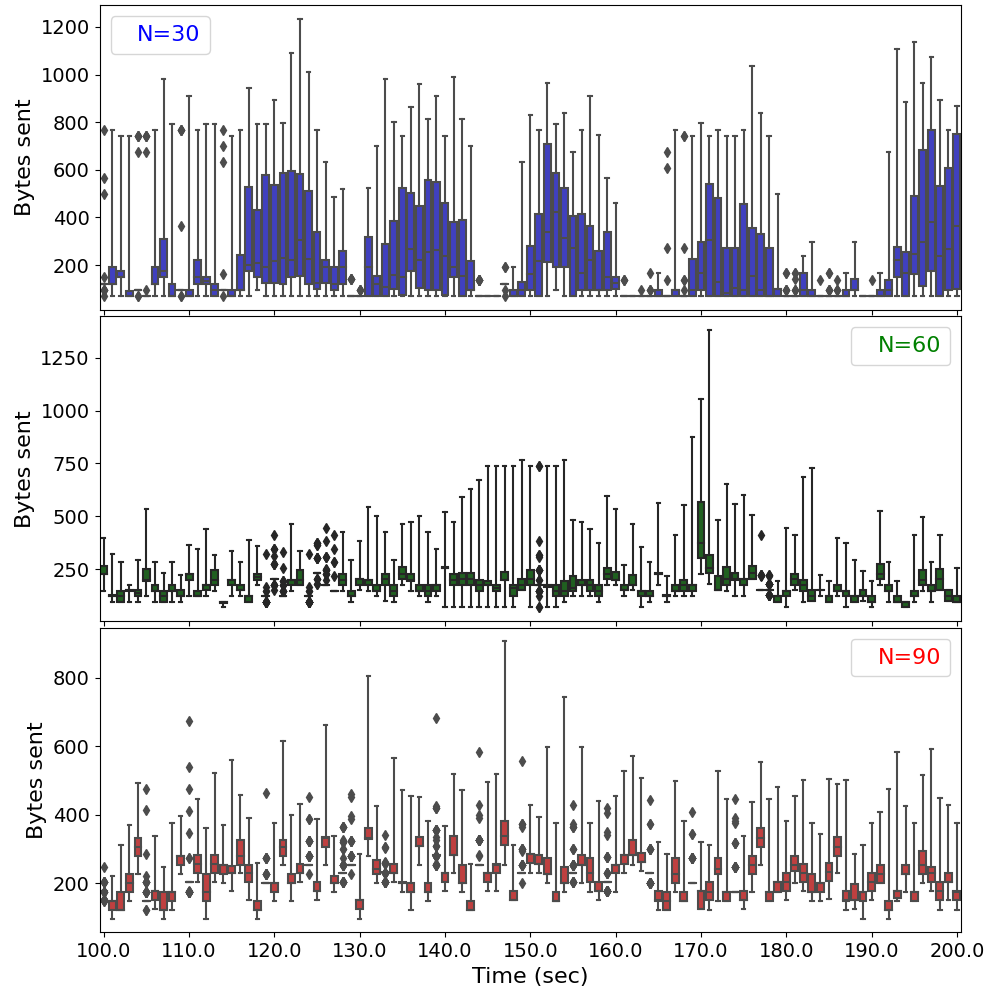}
    \caption{Bytes sent per second per robot over time for $V=6$ and $N \in \{30, 60, 90\}$.}
    \label{fig:bandwidthN}
\end{figure}

\textbf{Effect of the number of robots.} 
Figure~\ref{fig:bandwidthN} shows the bandwidth usage per robot across different values of $N$ in the configuration with the largest amount of data to be managed ($V=6$). Communication overhead decreases when $N$ increases because we keep the same shared memory parameters across values of $N$. This is consistent with the effect described when discussing the SwarmMesh dimensioning. 

\section{CONCLUSIONS} 
\label{sec:conclusions}
We proposed a method for the distributed storage and fusion of semantic annotations across a swarm of robots. We consider robots with limited memory, each running a pre-trained classifier to annotate objects in the environment. We show the costs of storing and reducing the uncertainty of a semantic map in terms of mapping performance, memory usage and communication overhead for users of this framework to make informed decisions on the trade-offs between the key performance metrics. In future work, we will study the aggregation of outputs of different types of classifiers to produce richer semantic maps. We will also assess the performance of our approach on real robots.


\section*{ACKNOWLEDGMENTS}
This work was funded by a grant from Amazon Robotics. This research was performed using computational resources supported by the Academic \& Research Computing group at Worcester Polytechnic Institute. 

\bibliographystyle{IEEEtran}
\bibliography{references}

\begin{thebibliography}{10}
\providecommand{\url}[1]{#1}
\csname url@samestyle\endcsname
\providecommand{\newblock}{\relax}
\providecommand{\bibinfo}[2]{#2}
\providecommand{\BIBentrySTDinterwordspacing}{\spaceskip=0pt\relax}
\providecommand{\BIBentryALTinterwordstretchfactor}{4}
\providecommand{\BIBentryALTinterwordspacing}{\spaceskip=\fontdimen2\font plus
\BIBentryALTinterwordstretchfactor\fontdimen3\font minus
  \fontdimen4\font\relax}
\providecommand{\BIBforeignlanguage}[2]{{%
\expandafter\ifx\csname l@#1\endcsname\relax
\typeout{** WARNING: IEEEtran.bst: No hyphenation pattern has been}%
\typeout{** loaded for the language `#1'. Using the pattern for}%
\typeout{** the default language instead.}%
\else
\language=\csname l@#1\endcsname
\fi
#2}}
\providecommand{\BIBdecl}{\relax}
\BIBdecl

\bibitem{valentini2016}
G.~Valentini, D.~Brambilla, H.~Hamann, and M.~Dorigo, ``{Collective Perception
  of Environmental Features in a Robot Swarm},'' in \emph{Swarm
  Intelligence}.\hskip 1em plus 0.5em minus 0.4em\relax Cham: Springer
  International Publishing, 2016, pp. 65--76.

\bibitem{ebert2020bayes}
J.~T. Ebert, M.~Gauci, F.~Mallmann-Trenn, and R.~Nagpal, ``Bayes bots:
  Collective bayesian decision-making in decentralized robot swarms,'' in
  \emph{2020 IEEE International Conference on Robotics and Automation
  (ICRA)}.\hskip 1em plus 0.5em minus 0.4em\relax IEEE, 2020, pp. 7186--7192.

\bibitem{riazuelo2015roboearth}
L.~Riazuelo, M.~Tenorth, D.~Di~Marco, M.~Salas, D.~G{\'a}lvez-L{\'o}pez,
  L.~M{\"o}senlechner, L.~Kunze, M.~Beetz, J.~D. Tard{\'o}s, L.~Montano
  \emph{et~al.}, ``Roboearth semantic mapping: A cloud enabled knowledge-based
  approach,'' \emph{IEEE Transactions on Automation Science and Engineering},
  vol.~12, no.~2, pp. 432--443, 2015.

\bibitem{khodayi2019distributed}
R.~Khodayi-mehr, Y.~Kantaros, and M.~M. Zavlanos, ``Distributed state
  estimation using intermittently connected robot networks,'' \emph{IEEE
  Transactions on Robotics}, vol.~35, no.~3, pp. 709--724, 2019.

\bibitem{atanasov2015decentralized}
N.~Atanasov, J.~Le~Ny, K.~Daniilidis, and G.~J. Pappas, ``Decentralized active
  information acquisition: Theory and application to multi-robot slam,'' in
  \emph{2015 IEEE International Conference on Robotics and Automation
  (ICRA)}.\hskip 1em plus 0.5em minus 0.4em\relax IEEE, 2015, pp. 4775--4782.

\bibitem{leung2012decentralized}
K.~Y. Leung, T.~D. Barfoot, and H.~H. Liu, ``Decentralized cooperative slam for
  sparsely-communicating robot networks: A centralized-equivalent approach,''
  \emph{Journal of Intelligent \& Robotic Systems}, vol.~66, no.~3, pp.
  321--342, 2012.

\bibitem{polikarEnsemble2012}
R.~Polikar, \emph{Ensemble Learning}, C.~Zhang and Y.~Ma, Eds.\hskip 1em plus
  0.5em minus 0.4em\relax Boston, MA: Springer US, 2012.

\bibitem{sunderhauf2018limits}
N.~S{\"u}nderhauf, O.~Brock, W.~Scheirer, R.~Hadsell, D.~Fox, J.~Leitner,
  B.~Upcroft, P.~Abbeel, W.~Burgard, M.~Milford \emph{et~al.}, ``The limits and
  potentials of deep learning for robotics,'' \emph{The International Journal
  of Robotics Research}, vol.~37, no. 4-5, pp. 405--420, 2018.

\bibitem{kuncheva2014combining}
L.~I. Kuncheva, \emph{Combining pattern classifiers: methods and
  algorithms}.\hskip 1em plus 0.5em minus 0.4em\relax John Wiley \& Sons, 2014.

\bibitem{uy2019revisiting}
M.~A. Uy, Q.-H. Pham, B.-S. Hua, T.~Nguyen, and S.-K. Yeung, ``Revisiting point
  cloud classification: A new benchmark dataset and classification model on
  real-world data,'' in \emph{Proceedings of the IEEE International Conference
  on Computer Vision}, 2019, pp. 1588--1597.

\bibitem{ValFerDor2017}
\BIBentryALTinterwordspacing
G.~Valentini, E.~Ferrante, and M.~Dorigo, ``The best-of-\emph{n} problem in
  robot swarms: {F}ormalization, state of the art, and novel perspectives,''
  \emph{Frontiers in Robotics and AI}, vol.~4, p.~9, 2017. [Online]. Available:
  \url{http://journal.frontiersin.org/article/10.3389/frobt.2017.00009}
\BIBentrySTDinterwordspacing

\bibitem{crosscombe2017robust}
M.~Crosscombe, J.~Lawry, S.~Hauert, and M.~Homer, ``Robust distributed
  decision-making in robot swarms: Exploiting a third truth state,'' in
  \emph{2017 IEEE/RSJ International Conference on Intelligent Robots and
  Systems (IROS)}.\hskip 1em plus 0.5em minus 0.4em\relax IEEE, 2017, pp.
  4326--4332.

\bibitem{mitra2019resilient}
A.~Mitra, J.~A. Richards, S.~Bagchi, and S.~Sundaram, ``Resilient distributed
  state estimation with mobile agents: overcoming byzantine adversaries,
  communication losses, and intermittent measurements,'' \emph{Autonomous
  Robots}, vol.~43, no.~3, pp. 743--768, 2019.

\bibitem{tahbaz-salehiConsensusRandomNetworks2006}
A.~{Tahbaz-Salehi} and A.~Jadbabaie, ``On consensus over random networks,'' in
  \emph{44th {{Annual Allerton Conference}}}.\hskip 1em plus 0.5em minus
  0.4em\relax {Citeseer}, 2006.

\bibitem{xiaoDistributedAverageConsensus2007}
L.~Xiao, S.~Boyd, and S.-J. Kim, ``\BIBforeignlanguage{en}{Distributed average
  consensus with least-mean-square deviation},''
  \emph{\BIBforeignlanguage{en}{Journal of Parallel and Distributed
  Computing}}, vol.~67, no.~1, pp. 33--46, Jan. 2007.

\bibitem{leblancResilientAsymptoticConsensus2013}
H.~J. Leblanc, H.~Zhang, X.~Koutsoukos, and S.~Sundaram, ``Resilient asymptotic
  consensus in robust networks,'' \emph{Ieee Journal on Selected Areas in
  Communications}, vol.~31, no.~4, 2013.

\bibitem{albani2017field}
D.~Albani, D.~Nardi, and V.~Trianni, ``Field coverage and weed mapping by uav
  swarms,'' in \emph{2017 IEEE/RSJ International Conference on Intelligent
  Robots and Systems (IROS)}.\hskip 1em plus 0.5em minus 0.4em\relax Ieee,
  2017, pp. 4319--4325.

\bibitem{lajoie2020door}
P.-Y. Lajoie, B.~Ramtoula, Y.~Chang, L.~Carlone, and G.~Beltrame, ``Door-slam:
  Distributed, online, and outlier resilient slam for robotic teams,''
  \emph{IEEE Robotics and Automation Letters}, vol.~5, no.~2, pp. 1656--1663,
  2020.

\bibitem{queralta2020collaborative}
J.~P. Queralta, J.~Taipalmaa, B.~C. Pullinen, V.~K. Sarker, T.~N. Gia,
  H.~Tenhunen, M.~Gabbouj, J.~Raitoharju, and T.~Westerlund, ``Collaborative
  multi-robot systems for search and rescue: Coordination and perception,''
  \emph{arXiv preprint arXiv:2008.12610}, 2020.

\bibitem{ruiz2017building}
J.-R. Ruiz-Sarmiento, C.~Galindo, and J.~Gonzalez-Jimenez, ``Building
  multiversal semantic maps for mobile robot operation,'' \emph{Knowledge-Based
  Systems}, vol. 119, pp. 257--272, 2017.

\bibitem{diffusion}
A.~Howard, M.~Mataric, and G.~Sukhatme, ``Mobile sensor network deployment
  using potential fields: A distributed, scalable solution to the area coverage
  problem,'' in \emph{DARS}, 2002.

\bibitem{mousavian20173d}
A.~Mousavian, D.~Anguelov, J.~Flynn, and J.~Kosecka, ``3d bounding box
  estimation using deep learning and geometry,'' in \emph{Proceedings of the
  IEEE Conference on Computer Vision and Pattern Recognition}, 2017, pp.
  7074--7082.

\bibitem{carrillo2014probabilistic}
H.~Carrillo, K.~H. Brodersen, and J.~A. Castellanos, ``Probabilistic
  performance evaluation for multiclass classification using the posterior
  balanced accuracy,'' in \emph{ROBOT2013: First Iberian Robotics
  Conference}.\hskip 1em plus 0.5em minus 0.4em\relax Springer, 2014, pp.
  347--361.

\bibitem{scenenn-3dv16}
B.-S. Hua, Q.-H. Pham, D.~T. Nguyen, M.-K. Tran, L.-F. Yu, and S.-K. Yeung,
  ``Scenenn: A scene meshes dataset with annotations,'' in \emph{International
  Conference on 3D Vision (3DV)}, 2016.

\bibitem{crainic2011efficient}
T.~G. Crainic, G.~Perboli, W.~Rei, and R.~Tadei, ``Efficient lower bounds and
  heuristics for the variable cost and size bin packing problem,''
  \emph{Computers \& Operations Research}, vol.~38, no.~11, pp. 1474--1482,
  2011.

\bibitem{majcherczyk2019swarmmesh}
N.~Majcherczyk and C.~Pinciroli, ``Swarm{M}esh: A distributed data structure
  for cooperative multi-robot applications,'' \emph{IEEE International
  Conference on Robotics and Automation}, 2020.

\bibitem{andrews1998theory}
G.~E. Andrews, \emph{The theory of partitions}.\hskip 1em plus 0.5em minus
  0.4em\relax Cambridge university press, 1998, no.~2.

\bibitem{Pinciroli:2012}
C.~Pinciroli, V.~Trianni, R.~O'Grady, G.~Pini, A.~Brutschy, M.~Brambilla,
  N.~Mathews, E.~Ferrante, G.~{Di Caro}, F.~Ducatelle, M.~Birattari, L.~M.
  Gambardella, and M.~Dorigo, ``{ARGoS}: a modular, parallel, multi-engine
  simulator for multi-robot systems,'' \emph{Swarm Intelligence}, vol.~6,
  no.~4, pp. 271--295, 2012.

\end{thebibliography}

\end{document}